\documentclass[letterpaper]{article} % DO NOT CHANGE THIS
\usepackage[preprint]{aaai2027}
\usepackage[hyphens]{url}  % DO NOT CHANGE THIS
\usepackage{graphicx} % DO NOT CHANGE THIS
\usepackage{natbib}  % DO NOT CHANGE THIS AND DO NOT ADD ANY OPTIONS TO IT
\usepackage{caption} % DO NOT CHANGE THIS AND DO NOT ADD ANY OPTIONS TO IT
\usepackage{amsmath}
\usepackage{amssymb}
\usepackage{booktabs}

\title{CEVA: Complementary Evidence Verification via Frame-Level Reorganization\\for Short-Video Fake News Detection}
\author{
    Wenshen Yang\textsuperscript{\rm 1},
    Junfeng Hao\textsuperscript{\rm 1},
    Jing Li\textsuperscript{\rm 1},
    Decheng Liu\textsuperscript{\rm 2},\\
    Yangkunfei\textsuperscript{\rm 3},
    Shuyu Jiang\textsuperscript{\rm 1},
    Haizhou Wang\textsuperscript{\rm 1}\corresponding
}
\affiliations{
    \textsuperscript{\rm 1}School of Cyber Science and Engineering, Sichuan University\\
    \textsuperscript{\rm 2}School of Cyber Engineering, Xidian University\\
    \textsuperscript{\rm 3}School of International Studies, Sichuan University\\
    whzh.nc@scu.edu.cn
}

\begin{document}

\maketitle

\begin{abstract}
Short-video platforms are now a common way for the public to follow unfolding events. Their rapid, visually persuasive format also makes fake news easier to spread. Unlike text--image posts, short-video news interleaves multiple modalities and changing scenes, making it difficult to identify which frames verify a claim. Existing detectors either pool sampled frames homogeneously or select them by text--frame similarity. Both overlook that aligned frames may repeat a misleading claim, whereas less dominant context may challenge it. We present \textbf{\textsc{CEVA}} (\textbf{C}omplementary \textbf{E}vidence \textbf{V}erification \textbf{A}pproach), to our knowledge the first short-video fake news detection framework to explicitly organize frames by their evidential roles before multimodal fusion. Its Dual-Branch Complementary Reorganization (DBCR) module separates explicit-alignment frames from candidate complementary-context frames, and cross-modal fusion with bidirectional contrastive regularization operates on the reorganized evidence. On FakeSV and FakeTT, CEVA achieves the best overall results among the compared methods. Under fixed frame budgets and backbones, DBCR outperforms uniform, random, and similarity-based selection and improves two structurally different detectors. Together, these results establish pre-fusion evidence organization as an effective and transferable component of short-video fake news detection.

\end{abstract}

\section{Introduction}

News increasingly reaches the public through short-video platforms, where visual, audio, textual, and social signals are woven into rapidly edited posts~\cite{buVideoSurvey2023,qiFakeSVMultimodalBenchmark2023,buFakingRecipeDetectingFake2024}. The same format gives fake news both reach and plausibility: misleading claims can be attached to persuasive footage and circulate farther and faster than verified information~\cite{shuFakeNewsSurvey2017,lazerScienceFakeNews2018,vosoughiSpread2018,zhouSurveyFakeNews2020}. Automated detection must therefore interpret heterogeneous evidence across the changing scenes and modalities of a short video.

\begin{figure}[t]
\centering
\includegraphics[width=\columnwidth]{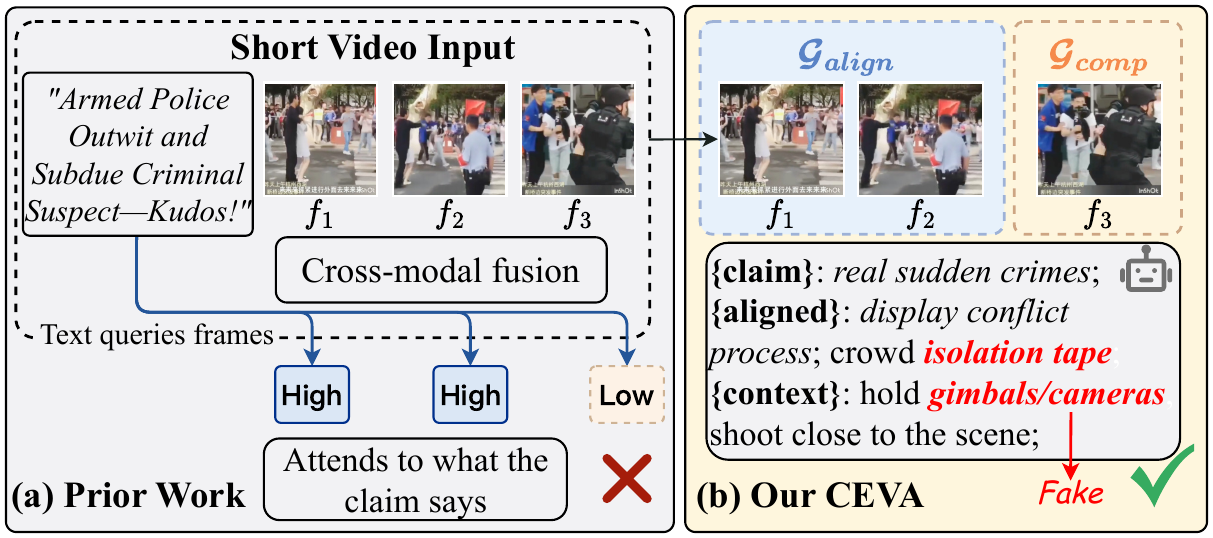}
\caption{Differences in frame-level evidence processing between prior multimodal methods and CEVA.}
\label{fig:motivation}
\end{figure}

Many fake news videos contain no altered pixels. Existing corpora document repurposed user-generated footage, misleading metadata, mismatched headlines, and genuine clips placed in misleading contexts~\cite{papadopoulouCorpus2019,palodMisleadingMetadata2019,sungVMH2023,liuCOVIDVTS2023}. Such a clip may match the reported event category while failing to verify its entity, place, time, or setting. Global modality agreement is therefore insufficient; detection must determine what individual frames contribute to the claim, as previewed in Figure~\ref{fig:motivation}.

Earlier multimodal fake news research studied recurrent fusion, event-invariant learning, and variational representations for text--image posts~\cite{jinMultimodalFusion2017,wangEANN2018,khattarMVAE2019}. Short-video detectors add caption guidance, content--social interaction, creative-process modeling, and event-neighbor reasoning~\cite{shangMultimodalMisinformationDetector2021a,qiFakeSVMultimodalBenchmark2023,buFakingRecipeDetectingFake2024,qiNEED2023}. More recent methods use counterfactual debiasing, adaptive fusion, denoising, or fine-grained consistency modeling~\cite{zong_text-guided_2025,shenMSAFNet2025,dengDenoisingEnhancedMultimodalDetection2026a,tian_exposing_2026}. Despite stronger multimodal interaction, these methods largely operate on an already sampled or encoded visual sequence.

\textbf{Sampled frames are commonly treated as a homogeneous evidence stream.} Fusion models aggregate frame features as one visual modality, although a video may alternate between the claimed action and brief contextual shots~\cite{qiFakeSVMultimodalBenchmark2023,zhangMFSVFND2025,zengMMVD2024}. Dominant action can then suppress diagnostic context.

\textbf{Verification utility extends beyond cross-modal relevance.} Caption guidance, similarity-aware fusion, and denoising emphasize claim-related content~\cite{shangMultimodalMisinformationDetector2021a,shenMSAFNet2025,dengDenoisingEnhancedMultimodalDetection2026a}. A closely aligned frame may reinforce a misleading narrative, while the least similar frames may be noise. Verification therefore requires claim-related frames with different evidential roles to be preserved before fusion.

Figure~\ref{fig:motivation} shows footage of an emergency-response drill near Broken Bridge at West Lake, reposted as ``Armed police officers outsmart and subdue a criminal! Bravo!'' The apprehension strongly matches the claim, but less prominent frames reveal close-range filming and barriers separating onlookers. Together, these cues point to an organized recording and call the claimed emergency into question. This case illustrates \emph{frame-level evidential heterogeneity}: frames from the same video can bear different relationships to the claim.

To test whether organizing these roles improves verification, we propose \textbf{\textsc{CEVA}} (\textbf{C}omplementary \textbf{E}vidence \textbf{V}erification \textbf{A}pproach). Its DBCR module combines shot coverage with each frame's similarity profile across five textual views, forming explicit-alignment ($\mathcal{G}_{\text{align}}$) and candidate complementary-context ($\mathcal{G}_{\text{comp}}$) groups before fusion. Controlled comparisons hold the frame budget, text streams, and backbone fixed, while plug-in experiments test DBCR on two structurally different detectors.

Our main contributions are:

\begin{itemize}
\item \textbf{We identify frame-level evidential heterogeneity as an overlooked source of verification evidence in short videos.} Claim-aligned frames and non-dominant contextual frames can play different diagnostic roles, challenging the use of text--frame similarity as the sole criterion for frame relevance.

\item \textbf{We operationalize this insight by reorganizing complementary frame evidence before multimodal fusion.} DBCR combines shot-aware sampling, multi-view semantic profiling, and profile-structure-based ranking to preserve both explicit-alignment and candidate complementary-context frames.

\item \textbf{We establish pre-fusion evidence organization as an effective and transferable design choice.} Controlled comparisons isolate DBCR under fixed frame budgets, backbones, and text streams, while plug-in evaluations on SV-FEND and FakingRecipe demonstrate gains across different fusion backbones.
\end{itemize}

\begin{figure*}[t]
\centering
\includegraphics[width=0.95\textwidth]{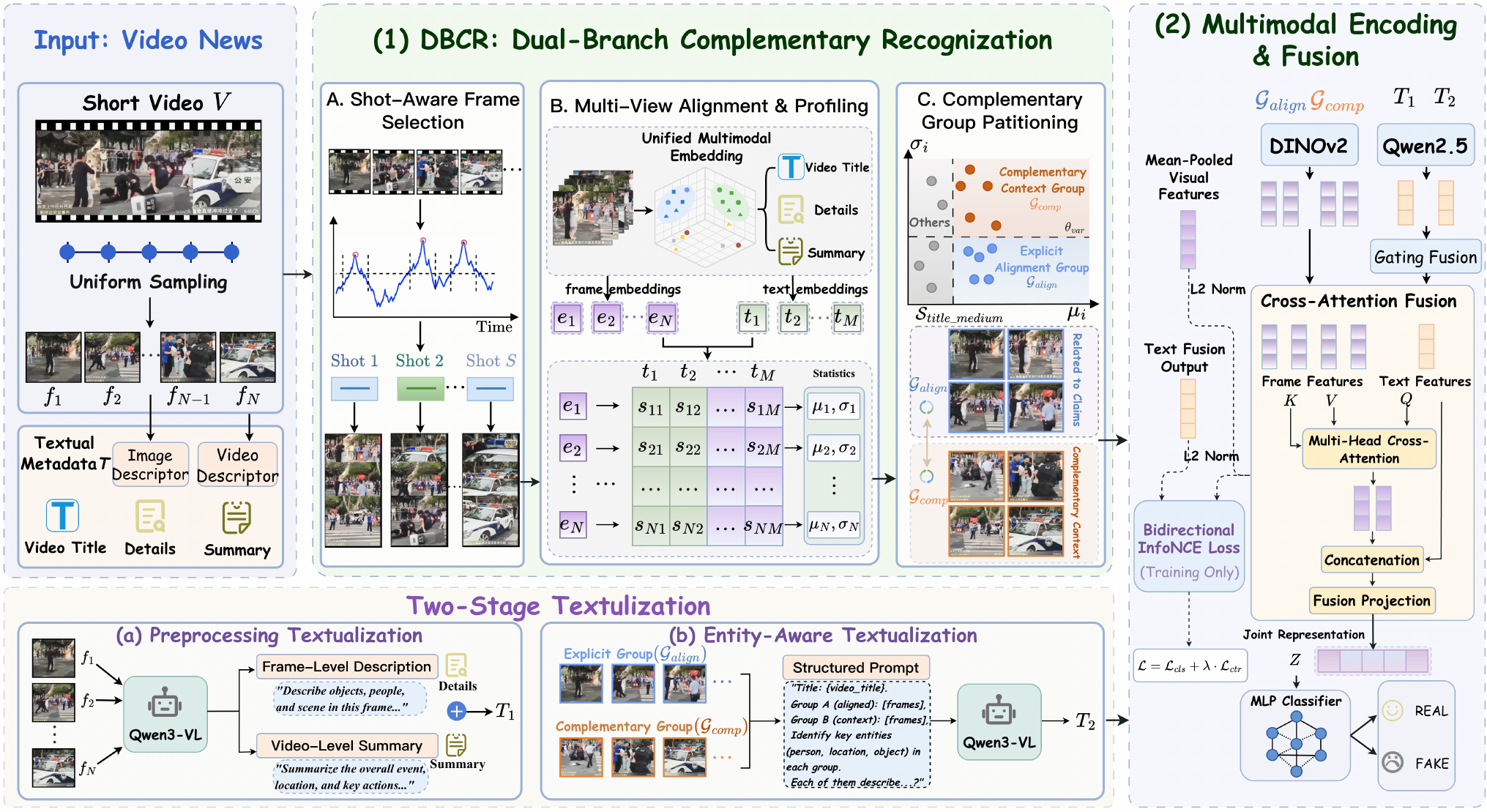}
\caption{Overall architecture of CEVA. DBCR reorganizes candidate frames before fusion, two-stage textualization constructs the global $T_1$ and entity-aware local $T_2$ streams, and bidirectional InfoNCE regularizes the fused representations during training.}
\label{fig:model}
\end{figure*}

\section{Related Work}

\subsection{Fake News Video Detection}

Fake news video detection has progressed from source and metadata verification~\cite{papadopoulouCorpus2019,palodMisleadingMetadata2019} to multimodal content modeling~\cite{buVideoSurvey2023}. Early systems combine video, language, and user signals~\cite{houAutomaticMedicalVideos2019,medinaSerranoCOVIDVideo2020}, while later work incorporates domain knowledge and comment relations~\cite{choiEffectiveFakeNews2022,choiUsingTopicModeling2021}.

Short-video benchmarks prompted architectures for densely edited posts. TikTec uses caption guidance~\cite{shangMultimodalMisinformationDetector2021a}; SV-FEND integrates content and social context~\cite{qiFakeSVMultimodalBenchmark2023}; NEED exploits event neighbors~\cite{qiNEED2023}; FakingRecipe models material-selection and editing preferences~\cite{buFakingRecipeDetectingFake2024}; and MMVD targets world bias~\cite{zengMMVD2024}. Related studies address health claims, misleading headlines, modality tampering, and opinion evolution~\cite{liuCOVIDVTS2023,sungVMH2023,wuInterpretableTampering2024,zongOpEvFake2024}.

Recent detectors increase granularity through synchronized comments, adaptive consistency, stronger fusion, and incomplete-modality learning~\cite{yinTSCMDFFM2025,shenMSAFNet2025,zhangMFSVFND2025,zengIMOL2025}. TGFC-SVFN uses text-guided counterfactual inference~\cite{zong_text-guided_2025}, MVA aggregates multiple evidence views~\cite{li_detecting_2026}, and MAGIC3 models token- and frame-level tri-modal consistency~\cite{tian_exposing_2026}. These methods reason over a sampled sequence; CEVA asks how that sequence should preserve frames with different claim relationships.

\subsection{LLM- and Knowledge-Enhanced Detection}

Large multimodal models introduce generated rationales and structured knowledge. ExMRD distills MLLM reasoning~\cite{hongFollowingCluesApproaching2025}; LVLM4FV, Fact-R1, and 3MFact organize multimodal reasoning or explanations~\cite{tahmasebiMultimodalMisinformationDetection2024,zhangFACTR1ExplainableVideo,niu2025pioneering}. FakeSV-VLM adapts a VLM for event checking~\cite{wang_fakesv-vlm_2025}, while FVDLM retrieves external news~\cite{han_enhancing_2026}. CIFE studies evidence-grounded explanations~\cite{chenCIFE2025}; other approaches refine structured knowledge or simulate creative processes~\cite{zengUnderstandRefineSummarize2025,buAgentAug2025}.

CEVA uses offline VLM textualization to organize evidence within the input video, then passes these representations to a trained detector for the final decision. Its reliance on within-video evidence distinguishes it from retrieval-based approaches, while the separation of frame organization and classification permits DBCR to be tested across detector backbones.

\section{Methodology}

\subsection{Problem Definition and Motivation}

Given $N$ sampled keyframes $\mathcal{V}=\{v_1,\ldots,v_N\}$, $v_i\in\mathbb{R}^{H\times W\times3}$, and text metadata $\mathcal{T}$, the task is to learn
\begin{equation}
f(\mathcal{V}, \mathcal{T}) \rightarrow \hat{y} \in \{0, 1\}
\label{eq:problem}
\end{equation}
where $\hat{y}=1$ denotes fake news. CEVA models the textual claim $\mathcal{T}$, an explicit-alignment group $\mathcal{G}_{\text{align}}$, and a complementary-context group $\mathcal{G}_{\text{comp}}$. The grouping uses similarity-based proxies to retain contextual candidates that homogeneous fusion or similarity truncation may underuse.

Training provides only the video-level label $y$; neither frame relevance nor evidence role is annotated. The two groups are therefore latent subsets constructed under a fixed visual budget:
\begin{equation}
\begin{aligned}
\mathcal{G}_{\text{align}},\mathcal{G}_{\text{comp}}&\subset\mathcal{V},
&\mathcal{G}_{\text{align}}\cap\mathcal{G}_{\text{comp}}&=\varnothing,\\
|\mathcal{G}_{\text{align}}|&=K,
&|\mathcal{G}_{\text{comp}}|&=K.
\end{aligned}
\label{eq:group_constraints}
\end{equation}
Equal cardinality fixes the information budget of each branch, and disjointness prevents the same frame from occupying both operational roles. These constraints preserve two structured candidate sets for downstream verification.

\subsection{Overall Framework}

Figure~\ref{fig:model} presents CEVA. A global text stream $T_1$ provides context for DBCR, which performs shot-aware sampling and partitions frames into explicit-alignment and complementary-context groups. Entity-aware textualization describes the reorganized groups as $T_2$. DINOv2 and Qwen-2.5 then encode the visual and textual streams; gated fusion and cross-attention integrate their evidence, while bidirectional InfoNCE regularizes paired representations.

\subsection{Dual-Branch Complementary Reorganization}

DBCR reorganizes $\mathcal{V}$ before fusion into a sequence $\mathcal{V}_{\text{seq}}$ with two evidence groups.

\subsubsection{Shot-Aware Candidate Frame Generation}

We segment the video into $M$ shots using $d_t=\|v_t-v_{t-1}\|_2$, with a boundary at $d_t>\mu_d+c\sigma_d$ and $c=1.5$. A shot $s_k$ receives a sampling weight based on its length and mean difference $\bar d_k$:
\begin{equation}
w_k=\frac{|s_k|}{|\mathcal{V}|}\left(1+\frac{\bar d_k}{\mu_d+\epsilon}\right),\qquad
n_k=N\frac{w_k}{\sum_{j=1}^{M}w_j}.
\label{eq:budget}
\end{equation}
Integer quotas are clipped to available frames; unused budget is filled from unsampled frames, preserving shot coverage while favoring dynamic segments.

\subsubsection{Multi-View Semantic Profiling}

Qwen3-VL-Embedding yields $\mathbf{f}_i=\text{VLE}(v_i)$ and text embeddings $\mathbf{t}_k=\text{VLE}(\mathcal{T}_k)$. Each frame receives a multi-view similarity profile:
\begin{equation}
\mathbf{p}_i = [\text{sim}(\mathbf{f}_i, \mathbf{t}_1), \ldots, \text{sim}(\mathbf{f}_i, \mathbf{t}_{K_v})]
\label{eq:profile}
\end{equation}
We extract alignment and dispersion statistics:
\begin{equation}
\alpha_i = \max_{k=1}^{K_v} \text{sim}_k(\mathbf{f}_i, \mathbf{t}), \qquad \beta_i = \text{std}(\mathbf{p}_i)
\label{eq:alpha_beta}
\end{equation}
Here $\alpha_i$ measures maximum alignment and $\beta_i$ measures cross-view variance.

\subsubsection{Profile-Structure-Based Partitioning}

The $K_v=5$ fields are title, claim, summary, details, and OCR, with $\mathbf{p}_i[k]=\cos(\mathbf{f}_i,\mathbf{t}_k)$. Title and OCR come from the post and frames; Qwen3-VL generates the other fields solely from these inputs and the video. No label, fact-check, or debunking information enters VLM generation.

After per-video min-max normalization, explicit alignment is scored as
\begin{equation}
s_i^{\text{align}} = \text{norm}(\text{sim}_{\text{title},i}) + \text{norm}(\alpha_i) + (1 - \text{norm}(\beta_i))
\label{eq:salign}
\end{equation}
which favors title alignment, cross-field consensus, and low dispersion.

Complementary evidence prioritizes dispersion while suppressing frames below median title similarity:
\begin{equation}
s_i^{\text{comp}} = \text{norm}(\beta_i) \cdot \mathbb{I}\bigl[\text{sim}_{\text{title},i} \geq \text{median}(\text{sim}_{\text{title}})\bigr]
\label{eq:scomp}
\end{equation}
where $\mathbb{I}[\cdot]$ is the indicator function.

The two groups are then formed by ranking:
\begin{align}
\mathcal{G}_{\text{align}} &= \text{Top-K}\bigl(\{s_i^{\text{align}}\}_{i=1}^{N}\bigr) \label{eq:galign}\\
\mathcal{G}_{\text{comp}} &= \text{Top-K}\bigl(\{s_i^{\text{comp}} \mid v_i \notin \mathcal{G}_{\text{align}}\}_{i=1}^{N}\bigr) \label{eq:gcomp}
\end{align}
Ranking yields $K$ disjoint frames per group without dataset-level thresholds. MMR~\cite{carbonell1998mmr} promotes diversity in $\mathcal{G}_{\text{align}}$, and a brightness/blur filter removes degenerate complementary frames. The groups are concatenated as $\mathcal{V}_{\text{seq}}=[\mathbf{v}_1^{(a)},\ldots,\mathbf{v}_{K}^{(a)},\mathbf{v}_1^{(c)},\ldots,\mathbf{v}_{K}^{(c)}]$.

\subsection{Multimodal Encoding}

\subsubsection{Text Encoding with Semantic Enhancement}

$T_1$ contains the video summary and details; $T_2$ contains entity-aware descriptions of the reorganized groups. Qwen-2.5-3B precomputes $\mathbf{e}_{T_1},\mathbf{e}_{T_2}\in\mathbb{R}^{2048}$. After projection $\mathbf{h}_{T_i}=\mathbf{W}_{\text{proj}}\mathbf{e}_{T_i}$ to $d=256$, learned gating combines them:
\begin{equation}
\mathbf{h}_{\text{text}} = \mathbf{g} \odot \mathbf{h}_{T_1} + (1 - \mathbf{g}) \odot \mathbf{h}_{T_2}
\label{eq:gate_fusion}
\end{equation}
\begin{equation}
\mathbf{g} = \sigma(\mathbf{W}_g[\mathbf{h}_{T_1};\mathbf{h}_{T_2}] + \mathbf{b}_g)
\label{eq:gate}
\end{equation}
The element-wise gate adjusts the contributions of global narrative and group-specific descriptions separately in each latent dimension. Offline precomputation avoids online LLM inference and adds approximately 8~KB per sample.

\subsubsection{Visual Encoding and Cross-Modal Fusion}

DINOv2 ViT-B/14 extracts $\mathbf{z}_i\in\mathbb{R}^{768}$. An inter-frame Transformer produces $\mathbf{Z}\in\mathbb{R}^{2K\times d}$, and $\mathbf{h}_{\text{vis}}=\text{MeanPool}(\mathbf{Z})$. The ordered sequence preserves group membership and within-group ranking, allowing frame interaction before temporal pooling.

Text-guided cross-attention selects visual evidence:
\begin{equation}
\mathbf{h}_{\text{cross}} = \text{MultiHeadAttn}(Q{=}\mathbf{h}_{\text{text}},\; K{=}\mathbf{Z},\; V{=}\mathbf{Z})
\label{eq:cross_attn}
\end{equation}
Here the fused text acts as the query over all $2K$ frame tokens, so aligned and complementary candidates remain separately accessible during evidence integration. The classifier receives
\begin{equation}
\mathbf{h}_{\text{joint}} = [\mathbf{h}_{\text{cross}};\mathbf{h}_{\text{text}};\mathbf{h}_{\text{vis}}]
\label{eq:joint}
\end{equation}

\subsection{Cross-Modal Contrastive Learning}

We use bidirectional InfoNCE for paired-instance regularization:
\begin{equation}
\mathcal{L}_{\text{contrast}} = \frac{1}{2}\bigl(\mathcal{L}_{\text{InfoNCE}}(\mathbf{h}_{\text{text}}, \mathbf{h}_{\text{vis}}) + \mathcal{L}_{\text{InfoNCE}}(\mathbf{h}_{\text{vis}}, \mathbf{h}_{\text{text}})\bigr)
\label{eq:contrast}
\end{equation}
with
\begin{equation}
\mathcal{L}_{\text{InfoNCE}}(\mathbf{u}, \mathbf{v}) = -\frac{1}{B}\sum_{i=1}^{B} \log \frac{\exp(\text{sim}(\mathbf{u}_i, \mathbf{v}_i) / \tau)}{\sum_{j=1}^{B} \exp(\text{sim}(\mathbf{u}_i, \mathbf{v}_j) / \tau)}
\label{eq:infonce}
\end{equation}

Text and video serve as queries in turn, producing different batch-wise normalization sets. Their average regularizes paired-instance correspondence; the classification loss provides veracity supervision.

\subsection{Training and Inference}

The training objective is
\begin{equation}
\mathcal{L} = \mathcal{L}_{\text{cls}} + \lambda \mathcal{L}_{\text{contrast}}
\label{eq:total_loss}
\end{equation}
where $\mathcal{L}_{\text{cls}}$ is binary cross-entropy. Qwen-2.5-3B, DINOv2, and Qwen3-VL-Embedding are frozen; only the projection, gating, inter-frame Transformer, cross-attention, and classifier layers are trained. We use AdamW ($1{\times}10^{-4}$ learning rate and weight decay) on one A800 GPU. Inference uses precomputed features and disables the contrastive branch.

\section{Experiments}

\subsection{Experimental Settings}

\textbf{Datasets.} We evaluate on FakeSV~\cite{qiFakeSVMultimodalBenchmark2023} and FakeTT~\cite{buFakingRecipeDetectingFake2024}. Table~\ref{tab:dataset_statistics} and Figure~\ref{fig:shot_distribution} summarize their composition and shot distributions. FakeSV contains nearly balanced Chinese videos from Douyin/Kuaishou, whereas FakeTT contains English TikTok videos with more fake than real samples. Their different languages, class ratios, and editing patterns test whether evidence organization transfers beyond one platform setting. We use an event-level 4:1 train/test split with no event overlap. Event separation prevents near-duplicate footage from making content memorization appear to be cross-event verification. All preprocessing is performed independently within each video and does not use test-set labels or global test statistics.

\begin{table}[!b]
\centering
\setlength{\tabcolsep}{4pt}
\begin{tabular}{lrrrr}
\toprule
Dataset & Videos (F/R) & Duration & Shots \\
\midrule
FakeSV & 3,624 (1,810/1,814) & 39.88 & 7.70 \\
FakeTT & 1,991 (1,172/819) & 47.69 & 8.10 \\
\bottomrule
\end{tabular}
\caption{Dataset statistics. Duration and shots are per-video averages (seconds/count); F/R denotes fake/real.}
\label{tab:dataset_statistics}
\includegraphics[width=0.96\columnwidth]{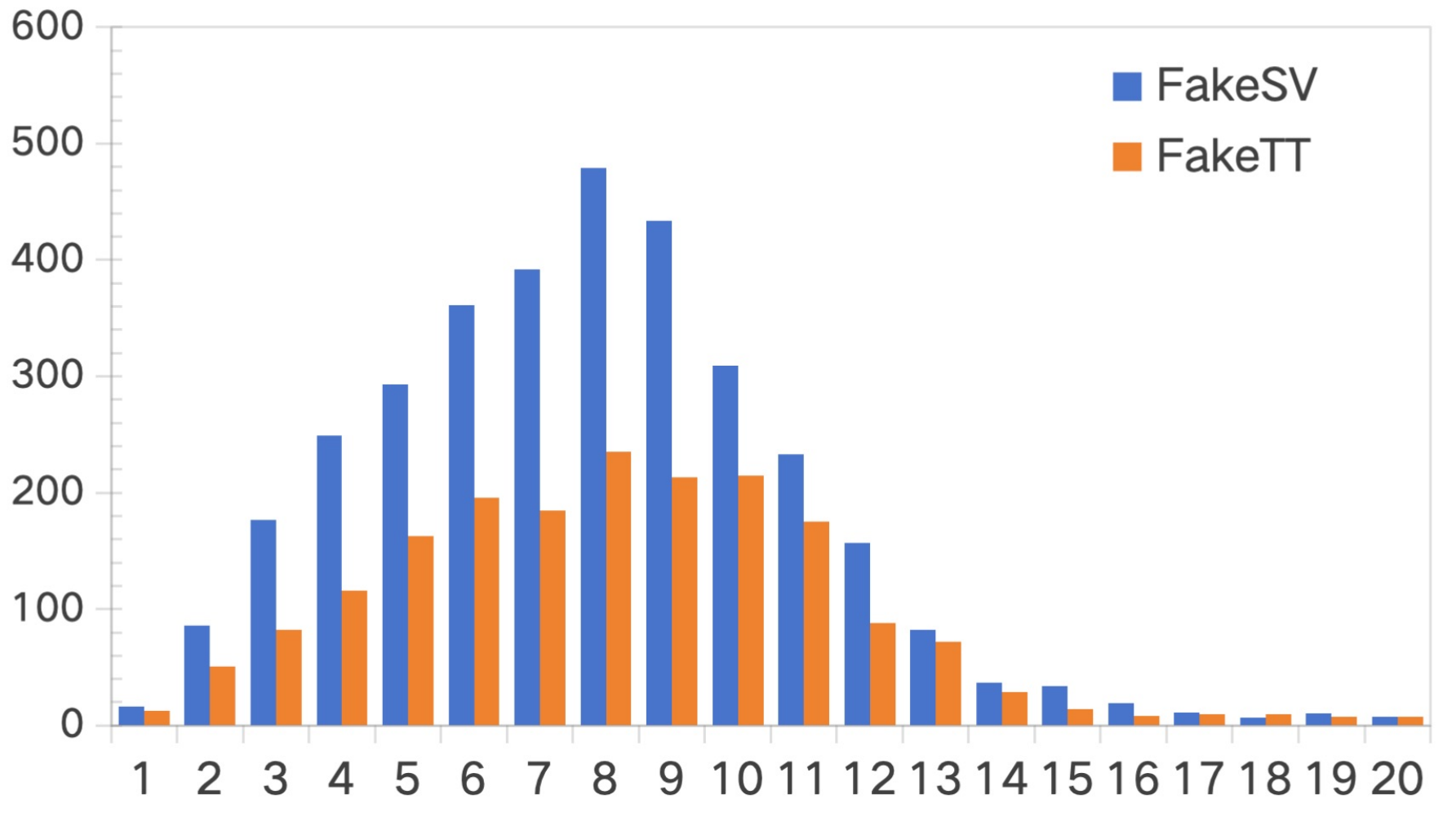}
\captionof{figure}{Shot-count distributions in FakeSV and FakeTT.}
\label{fig:shot_distribution}
\end{table}

\begin{table*}[t]
\centering
\setlength{\tabcolsep}{7pt}
\begin{tabular}{lrrrr@{\hskip 16pt}rrrr}
\toprule
& \multicolumn{4}{c}{FakeSV} & \multicolumn{4}{c}{FakeTT} \\
\cmidrule(lr){2-5} \cmidrule(lr){6-9}
Method & Acc & M-F1 & M-Prec & M-Rec & Acc & M-F1 & M-Prec & M-Rec \\
\midrule
BERT & 76.89 & 76.83 & 77.21 & 76.90 & 71.24 & 69.31 & 68.98 & 70.85 \\
ViT & 71.22 & 71.04 & 71.04 & 71.33 & 65.55 & 64.39 & 65.17 & 67.11 \\
\midrule
TikTec & 73.43 & 73.26 & 73.23 & 73.54 & 66.22 & 65.08 & 65.84 & 67.87 \\
FANVM & 79.52 & 78.81 & 79.81 & 78.46 & 71.57 & 70.21 & 70.22 & 72.63 \\
SV-FEND & 80.29 & 80.03 & 80.70 & 80.34 & 77.62 & 77.54 & 77.56 & 77.53 \\
FakingRecipe & 84.69 & 84.30 & 84.80 & 84.01 & 79.53 & 77.42 & 77.01 & 78.29 \\
D$^2$et & 85.79 & 85.39 & 86.08 & 85.06 & 83.95 & 82.31 & 81.90 & 83.16 \\
ExMRD & \underline{86.90} & \underline{86.52} & \underline{87.31} & \underline{86.13} & \underline{84.28} & \underline{83.13} & \underline{82.27} & \textbf{85.19} \\
\midrule
LLaVA-OV & 57.54 & 50.71 & 61.57 & 55.94 & 46.93 & 46.70 & 53.55 & 53.24 \\
Qwen2-VL & 53.85 & 53.72 & 54.29 & 54.20 & 53.17 & 52.71 & 56.27 & 57.65 \\
Qwen2.5-VL-7B & 55.60 & 55.50 & 55.70 & 55.60 & 54.90 & 54.00 & 54.10 & 54.00 \\
GPT-4o-mini & 66.42 & 65.88 & 65.90 & 65.87 & 57.85 & 57.78 & 62.91 & 63.65 \\
\midrule
\textbf{CEVA (Ours)} & \textbf{87.98} & \textbf{87.96} & \textbf{87.96} & \textbf{87.97} & \textbf{86.67} & \textbf{85.58} & \textbf{86.68} & \underline{84.86} \\
\bottomrule
\end{tabular}
\caption{Overall performance on FakeSV and FakeTT (\%). Best and second-best results are bold and underlined.}
\label{tab:main}
\end{table*}

\textbf{Baselines.} We compare against three categories. \emph{Unimodal}: BERT~\cite{devlinBERTPretrainingDeep2019} (title only) and ViT~\cite{dosovitskiyImageWorth16x162021} (keyframes only). \emph{Multimodal}: TikTec~\cite{shangMultimodalMisinformationDetector2021a}, FANVM~\cite{choiUsingTopicModeling2021}, SV-FEND~\cite{qiFakeSVMultimodalBenchmark2023}, FakingRecipe~\cite{buFakingRecipeDetectingFake2024}, D$^2$et~\cite{dengDenoisingEnhancedMultimodalDetection2026a}, and ExMRD~\cite{hongFollowingCluesApproaching2025}. \emph{MLLMs (zero-shot)}: GPT-4o-mini, LLaVA-OV, Qwen2-VL, and Qwen2.5-VL-7B.

\textbf{Evaluation.} We report Accuracy (Acc) together with macro-averaged F1 (M-F1), Precision (M-Prec), and Recall (M-Rec). Class-specific precision, recall, and F1 are additionally reported for the contrastive-objective comparison. We set $\lambda=0.25$, DBCR $K{=}6$, the maximum number of candidate frames to 32, and the batch size to 32, and optimize the trainable modules with AdamW. Controlled variants reuse the same candidate pool and cached multimodal features, changing only the component named in each comparison. All runs use seed 2025 and stop after five epochs without validation improvement.

\subsection{Main Results}

Table~\ref{tab:main} presents the overall comparison. CEVA records the highest accuracy among the compared results: 87.98\% on FakeSV and 86.67\% on FakeTT, 1.08 and 2.39 percentage points higher than ExMRD, respectively. On FakeTT, ExMRD obtains slightly higher macro-recall (85.19\% versus 84.86\%), while CEVA gains 4.41~pp in macro-precision and 2.45~pp in macro-F1. CEVA's advantage on this dataset thus lies in higher macro-precision and macro-F1, alongside a small decrease in macro-recall.

Multimodal methods outperform the unimodal baselines in Table~\ref{tab:main}, showing the benefit of combining titles and keyframes for this task. The reported zero-shot MLLM results remain well below trained detectors: GPT-4o-mini reaches 66.42\% on FakeSV and 57.85\% on FakeTT, while Qwen2.5-VL-7B obtains 55.60\% and 54.90\%. These results favor task-specific training over zero-shot prediction with broadly pretrained multimodal models. Table~\ref{tab:main} compares complete systems with different backbones and preprocessing pipelines. The ablation, controlled selection, and plug-in experiments below examine the contribution of frame reorganization more directly.

\subsection{Ablation Study}

Table~\ref{tab:ablation} reports the ablation results on both datasets.

\begin{table}[t]
\centering
\setlength{\tabcolsep}{2.8pt}
\begin{tabular}{lccc}
\toprule
Configuration & Acc & M-F1 & M-Rec \\
\midrule
CEVA (Full) & \textbf{87.98/86.67} & \textbf{87.96/85.58} & \textbf{87.97/84.86} \\
\midrule
w/o $T_1$ & 83.71/82.40 & 83.44/81.35 & 83.39/81.27 \\
w/o DBCR & 84.93/83.73 & 84.89/82.64 & 84.86/82.35 \\
w/o $\mathcal{G}_{\text{comp}}$ + $T_2$ & 84.48/83.47 & 84.40/82.38 & 84.33/82.13 \\
w/o $T_2$ & 85.69/84.27 & 85.66/83.38 & 85.63/82.82 \\
w/o Cross-Attn. & 84.17/85.33 & 84.15/84.83 & 84.13/84.43 \\
w/o $\mathcal{L}_{\text{contrast}}$ & 84.78/85.07 & 84.71/83.90 & 84.86/83.30 \\
\bottomrule
\end{tabular}
\caption{Ablation results (\%); entries are FakeSV/FakeTT.}
\label{tab:ablation}
\end{table}

Removing $T_1$ causes the largest drop (4.27~pp on both datasets), showing the importance of global declarative content from titles, descriptions, and creator-added on-screen text. The entity-aware local descriptions in $T_2$ are complementary: removing only $T_2$ reduces accuracy by 2.29/2.40~pp on FakeSV/FakeTT.

The w/o DBCR variant replaces the reorganized sequence with uniform keyframes and removes DBCR-dependent $T_2$, reducing accuracy by 3.05/2.94~pp. This joint ablation measures the combined effect of frame selection and its dependent text stream. Removing $\mathcal{G}_{\text{comp}}$ and $T_2$ costs 3.50/3.20~pp. With $T_2$ absent in both cases, uniform sampling is 0.45/0.26~pp better than retaining only $\mathcal{G}_{\text{align}}$. Alignment-only selection can be worse than no similarity filtering, although classification results do not establish the semantic purity of $\mathcal{G}_{\text{comp}}$.

Removing cross-attention costs 3.81/1.34~pp, while removing contrastive regularization costs 3.20/1.60~pp. Both components improve performance on the two datasets, with larger gains on FakeSV.

\begin{table*}[!t]
\centering
\setlength{\tabcolsep}{4pt}
\begin{tabular}{lcccccccc}
\toprule
\multicolumn{9}{l}{\textbf{(a) Frame-selection strategy}} \\
& & & \multicolumn{3}{c}{Fake} & \multicolumn{3}{c}{Real} \\
\cmidrule(lr){4-6} \cmidrule(lr){7-9}
Strategy & Acc & M-F1 & Prec & Rec & F1 & Prec & Rec & F1 \\
\midrule
Uniform & 87.37 & 87.32 & 89.04 & 84.28 & 86.59 & 85.96 & 90.27 & 88.06 \\
Random & 84.02 & 83.99 & 85.11 & 81.68 & 83.36 & 83.05 & 86.27 & 84.63 \\
Title-Top & 86.45 & 86.35 & \textbf{90.46} & 80.50 & 85.19 & 83.42 & \textbf{92.04} & 87.52 \\
Title-Bottom & 82.95 & 82.90 & 83.88 & 80.19 & 81.99 & 82.15 & 85.55 & 83.82 \\
Dispersion-Top & 85.24 & 85.17 & 87.46 & 81.13 & 84.18 & 83.43 & 89.09 & 86.16 \\
MMR & 83.26 & 83.23 & 83.12 & 82.08 & 82.59 & 83.38 & 84.37 & 83.87 \\
\textbf{CEVA (DBCR)} & \textbf{87.98} & \textbf{87.96} & 87.46 & \textbf{87.74} & \textbf{87.60} & \textbf{88.46} & 88.20 & \textbf{88.33} \\
\midrule
\multicolumn{9}{l}{\textbf{(b) Contrastive objective}} \\
& & & \multicolumn{3}{c}{Fake} & \multicolumn{3}{c}{Real} \\
\cmidrule(lr){4-6} \cmidrule(lr){7-9}
Objective & Acc & M-F1 & Prec & Rec & F1 & Prec & Rec & F1 \\
\midrule
None & 84.78 & 84.71 & 87.07 & 80.50 & 83.66 & 82.92 & 88.79 & 85.75 \\
T$\rightarrow$V & 85.84 & 85.83 & 85.49 & 85.22 & 85.35 & 86.18 & 86.43 & 86.30 \\
V$\rightarrow$T & 86.45 & 86.33 & \textbf{91.34} & 79.56 & 85.04 & 82.89 & \textbf{92.92} & 87.62 \\
Real-only & 85.69 & 85.61 & 88.89 & 80.50 & 84.49 & 83.20 & 90.56 & 86.72 \\
Label-cond. & 85.84 & 85.83 & 85.05 & 85.85 & 85.45 & 86.61 & 85.84 & 86.22 \\
\textbf{CEVA (Bi-dir.)} & \textbf{87.98} & \textbf{87.96} & 87.46 & \textbf{87.74} & \textbf{87.60} & \textbf{88.46} & 88.20 & \textbf{88.33} \\
\bottomrule
\end{tabular}
\caption{Controlled comparisons on FakeSV (\%). (a) Frame-selection strategies. (b) Contrastive objectives, where T$\rightarrow$V and V$\rightarrow$T denote text-to-video and video-to-text. CEVA uses DBCR with the bidirectional objective, so its full-model row is identical across panels. Column maxima are bold.}
\label{tab:frame_selection}
\label{tab:contrastive_objectives}
\end{table*}

\begin{table}[!t]
\centering
\begin{tabular}{lccc}
\toprule
Configuration & Acc & M-F1 & M-Rec \\
\midrule
\multicolumn{4}{c}{SV-FEND on FakeSV} \\
\midrule
Baseline & 80.29 & 80.03 & 80.34 \\
+ DBCR & \textbf{84.08} & \textbf{83.98} & \textbf{84.02} \\
\midrule
\multicolumn{4}{c}{FakingRecipe on FakeTT} \\
\midrule
Baseline & 79.53 & 77.42 & 78.29 \\
+ DBCR & \textbf{81.61} & \textbf{79.86} & \textbf{80.89} \\
\bottomrule
\end{tabular}
\caption{DBCR plug-in results on the native dataset of each baseline (\%).}
\label{tab:transfer}
\end{table}

\subsection{Further Analysis}

\subsubsection{Controlled Frame-Selection Comparison}

Table~\ref{tab:frame_selection} isolates frame selection on FakeSV. All variants use the same 32-frame candidate pool, 12-frame input budget, backbone, $T_1/T_2$ streams, data split, and training configuration; only the selection rule changes.

DBCR outperforms Uniform by 0.61~pp in accuracy and 0.64~pp in macro-F1 while keeping all textual information fixed. Without $T_2$, the same comparison favors DBCR by 0.76~pp (w/o $T_2$ versus w/o DBCR in Table~\ref{tab:ablation}). Adding $T_2$ improves Uniform and DBCR by 2.44 and 2.29~pp, respectively. The interaction is only $-0.15$~pp, so frame reorganization and local textualization contribute largely independently. Uniform remains the strongest single-rule alternative, indicating that broad temporal coverage is already valuable; DBCR improves on that coverage by assigning different roles to the retained frames.

Class-specific metrics expose a failure mode of similarity-only selection. Title-Top attains 90.46\% fake precision but only 80.50\% fake recall, while its real recall rises to 92.04\%; strongly claim-aligned videos are more often classified as real. Relative to Uniform, DBCR raises fake recall by 3.46~pp and fake F1 by 1.01~pp, while also improving real F1 by 0.27~pp. Random and MMR lose 3.96 and 4.72~pp in accuracy, respectively, showing that stochastic coverage or visual diversity alone is insufficient. Title-Bottom and Dispersion-Top likewise underperform. These comparisons favor DBCR's combined use of alignment, cross-field dispersion, and branch-specific organization over selection based on either score extreme.

\subsubsection{DBCR Plug-in Evaluation}

Table~\ref{tab:transfer} evaluates DBCR with two structurally different detectors on their native datasets: SV-FEND on FakeSV and FakingRecipe on FakeTT. Each detector is retained while DBCR reorganizes its frame input.

DBCR improves SV-FEND by 3.79/3.95/3.68~pp and FakingRecipe by 2.08/2.44/2.60~pp in accuracy/macro-F1/macro-recall. SV-FEND represents conventional cross-modal fusion, whereas FakingRecipe models material selection and editing patterns. Both detectors benefit from reorganized frame input while retaining their original architectures. The smaller gain on FakingRecipe may reflect partial functional overlap; however, the different evaluation datasets also limit direct comparison of the gain magnitudes.

\subsubsection{Contrastive Objective Comparison}

Table~\ref{tab:contrastive_objectives} compares alternative objectives on FakeSV with the architecture and split fixed. ``Real-only'' applies bidirectional InfoNCE only to real samples; label-conditioned treats samples with the same label as positives.

Bidirectional InfoNCE improves accuracy/macro-F1 by 3.20/3.25~pp over no contrastive loss and balances fake/real recall (87.74/88.20\%). Video-to-text is strongly asymmetric (79.56/92.92\%), while text-to-video, real-only, and label-conditioned objectives also underperform. This balance matters because the frame-selection controls show that one-sided alignment can improve recognition of one class while suppressing fake recall. The bidirectional objective achieves higher overall performance alongside near-equal class recalls. Its two directions use complementary batch-wise normalization sets to regularize instance correspondence, with veracity supervision supplied by the classification loss.

\begin{figure*}[!t]
\centering
\includegraphics[width=\textwidth]{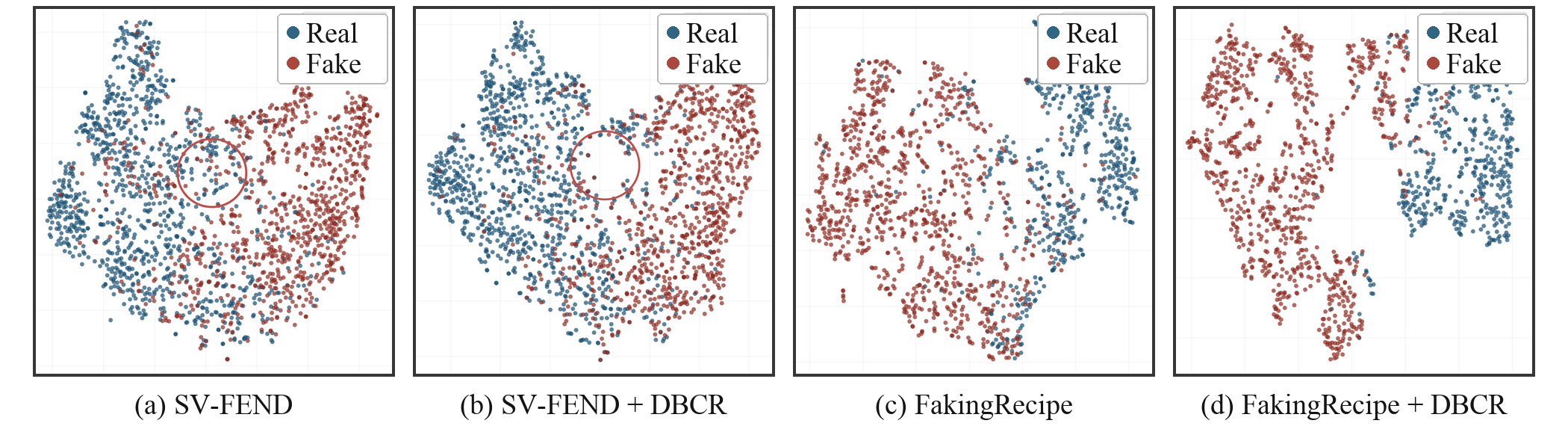}
\caption{t-SNE projections for the plug-in evaluations. DBCR reduces class overlap for SV-FEND on FakeSV and FakingRecipe on FakeTT. Coordinates are omitted because t-SNE axes have no semantic scale; blue/red denote real/fake samples.}
\label{fig:tsne_plugin}
\end{figure*}

\subsubsection{t-SNE Visualization}

Figure~\ref{fig:tsne_plugin} shows cleaner fake/real separation for both detector pairs after frame reorganization, consistent with the gains in Table~\ref{tab:transfer}. These video-level projections illustrate the geometry of learned representations. The semantics of the selected frame groups require separate assessment.

\subsubsection{Effect of Contrastive Weight $\lambda$}

\begin{figure}[!t]
\centering
\includegraphics[width=0.92\columnwidth]{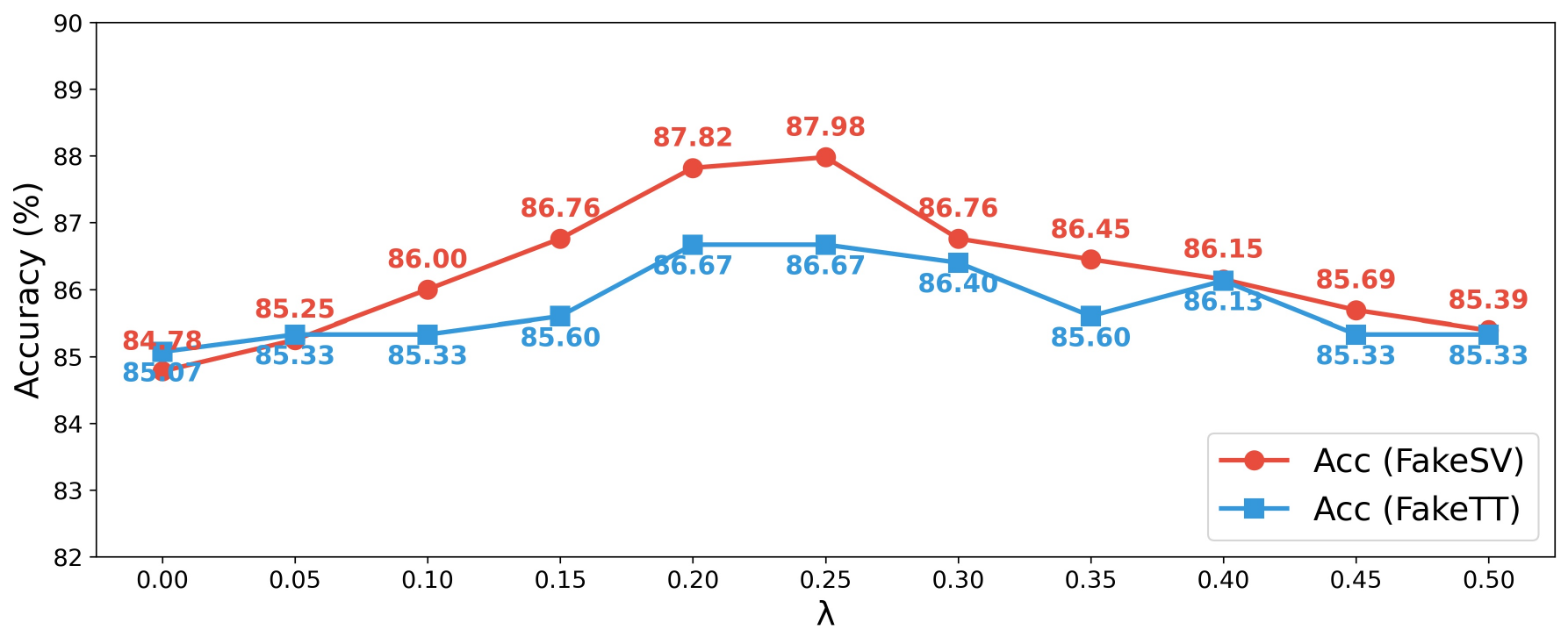}
\caption{Accuracy sensitivity to the contrastive loss weight $\lambda$ on FakeSV and FakeTT.}
\label{fig:lambda}
\end{figure}

Figure~\ref{fig:lambda} examines the sensitivity to the contrastive weight.

Accuracy peaks at $\lambda{=}0.25$ on FakeSV and across $\lambda\in[0.20,0.25]$ on FakeTT, then declines at larger weights. The shared trend supports using contrastive learning as a secondary regularizer.

\subsubsection{Case Study}

\begin{figure}[!t]
\centering
\includegraphics[width=\columnwidth]{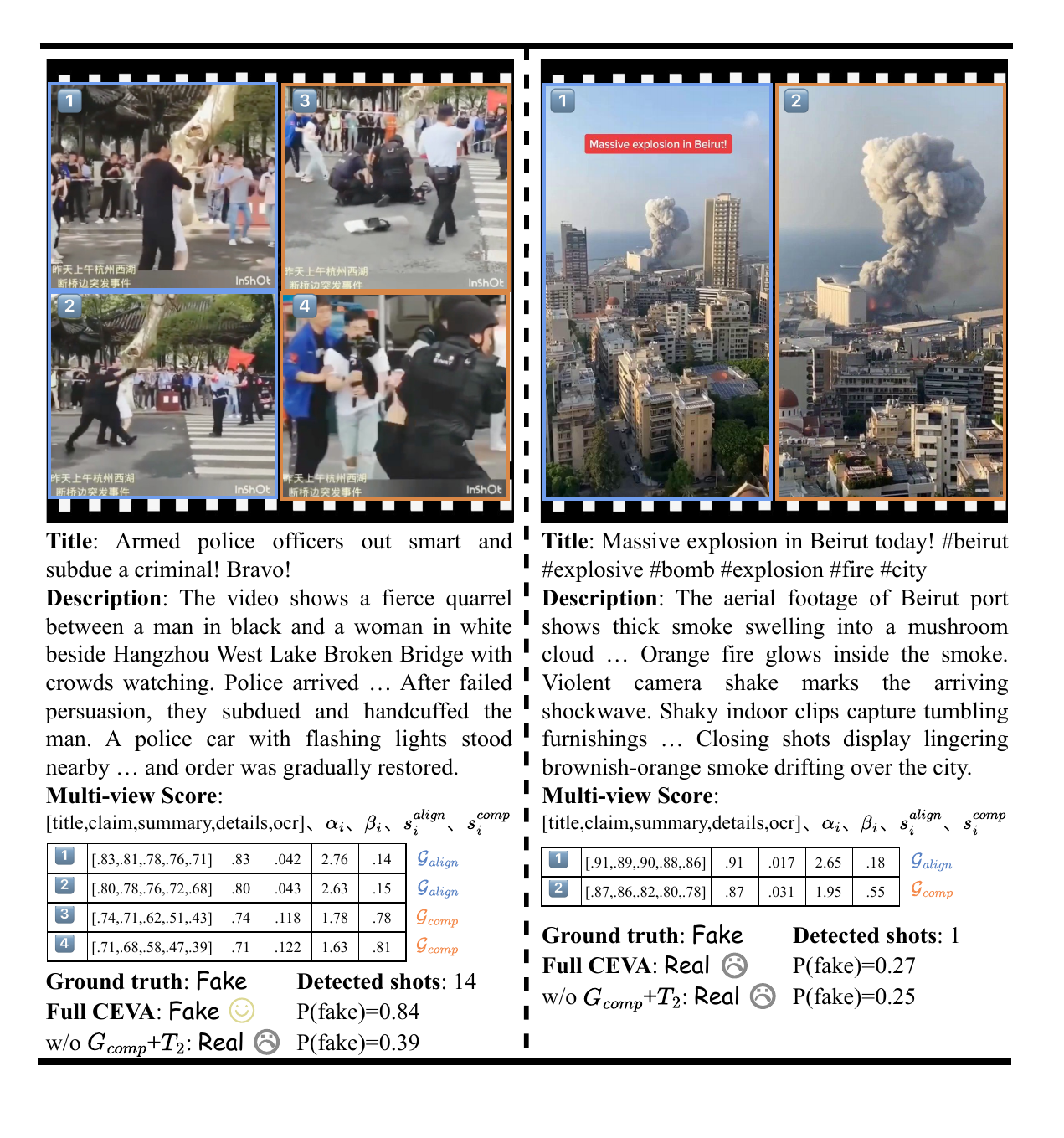}
\caption{Two fake news cases. (a) Left: CEVA identifies a 14-shot drill from aligned action and filming-context cues. (b) Right: CEVA misses an entity replacement in single-shot, title-aligned footage. Blue/orange denote $\mathcal{G}_{\text{align}}/\mathcal{G}_{\text{comp}}$.}
\label{fig:case_study}
\end{figure}

Figure~\ref{fig:case_study} returns to the West Lake example in Figure~\ref{fig:motivation} and contrasts it with a false negative.

In (a), aligned apprehension frames score 2.76/2.63, while filming-context frames receive complementary scores of 0.78/0.81. Adding $\mathcal{G}_{\text{comp}}$ and $T_2$ changes $P(\mathrm{fake})$ from 0.39 to 0.84. In (b), the one-shot explosion remains title-aligned throughout; an entity replacement leaves no visible contextual contradiction, and CEVA predicts real ($0.27$). DBCR returns $K$ candidates per branch, so $\mathcal{G}_{\text{comp}}$ contains the highest-ranked available candidates even when the video provides no distinct contextual evidence. Together, these cases show how CEVA's benefit depends on contextual variation within a video: filming cues can help identify the drill, while resolving the replaced entity in the explosion clip requires information beyond the available frames.

\FloatBarrier
\section{Conclusion}

In this work, we introduced CEVA to organize frame evidence before multimodal fusion. DBCR retains claim-aligned and candidate complementary-context frames through shot-aware sampling and profile-based partitioning. Controlled and plug-in experiments show that this organization improves class balance and transfers across detector architectures. The results highlight the importance of selecting frames according to their contribution to claim verification, with textual similarity serving as one aspect of relevance.

\bibliography{references}

\end{document}